%% file: main.tex
\documentclass[letterpaper, 10 pt, journal, twoside]{IEEEtran}  

\IEEEoverridecommandlockouts                              

\usepackage{ral}

\title{
Multi-Task Reinforcement Learning for Quadrotors
}
\author{Jiaxu Xing, Ismail Geles, Yunlong Song, Elie Aljalbout, and Davide Scaramuzza
    \thanks{
    Manuscript received: September 3, 2024; Revised November 6, 2024; Accepted November 26, 2024.

    This paper was recommended for publication by Editor Giuseppe Loianno upon evaluation of the Associate Editor and Reviewers' comments.
    
    The authors are with the Robotics and Perception Group, Department of Informatics, University of Zurich, and Department of Neuroinformatics, University of Zurich and ETH Zurich, Switzerland (\protect\url{http://rpg.ifi.uzh.ch}). This work was supported by the European Union’s Horizon Europe Research and Innovation Programme under grant agreement No. 101120732 (AUTOASSESS) and the European Research Council (ERC) under grant agreement No. 864042 (AGILEFLIGHT).}
    \thanks{Digital Object Identifier (DOI): see top of this page.}
}
\begin{document}
\maketitle

\input{sections/abstract}
\input{sections/introduction}

\input{sections/related_work}
\input{sections/methodology}

\input{sections/experiments}
\input{sections/discusssion}
\input{sections/conclusions}
\balance
\bibliographystyle{ieeeTran}
\bibliography{ral}

\end{document}

%% file: sections/abstract.tex
\begin{abstract}
Reinforcement learning (RL) has shown great effectiveness in quadrotor control, enabling specialized policies to develop even human-champion-level performance in single-task scenarios.
However, these specialized policies often struggle with novel tasks, requiring a complete retraining of the policy from scratch.
To address this limitation, this paper presents a novel multi-task reinforcement learning (MTRL) framework tailored for quadrotor control, leveraging the shared physical dynamics of the platform to enhance sample efficiency and task performance. 
By employing a multi-critic architecture and shared task encoders, our framework facilitates knowledge transfer across tasks, enabling a single policy to execute diverse maneuvers, including high-speed stabilization, velocity tracking, and autonomous racing. 
Our experimental results, validated both in simulation and real-world scenarios, demonstrate that our framework outperforms baseline approaches in terms of sample efficiency and overall task performance.

\noindent Video: \url{https://youtu.be/HfK9UT1OVnY}.
\end{abstract}

%% file: sections/introduction.tex
\section{Introduction}
\input{figures/eyecatcher_vertical}

\IEEEPARstart{R}{eal} world quadrotor applications typically involve multiple tasks and skills.
For example, in search and rescue scenarios or inspection, quadrotors are required to perform a range of specific tasks within a single mission, such as tracking moving objects, maintaining stable hover positions, and precisely following designated paths or targets.
To meet these diverse demands, a generalist control policy that can effectively manage these tasks can greatly enhance the versatility and adaptability of quadrotors, making them more effective in real-world applications.
However, developing a generalist controller for multiple tasks is a challenging problem
since different tasks often have different objectives and state spaces. For instance, in quadrotor control, the objective for hovering is to stabilize the vehicle by reducing its velocity to zero, whereas quadrotor racing requires maximizing speed while avoiding collisions with gates. These tasks inherently conflict with their goals and demand different observations and strategies. 

In this work, we tackle the multi-task quadrotor control problem using deep reinforcement learning (RL), which offers the advantage of automatically optimizing parametric controllers through trial and error. 
RL is particularly effective in handling highly non-linear dynamical systems and non-convex, non-differentiable objectives\textemdash challenges that are typically difficult for conventional optimization-based methods such as Model Predictive Control (MPC)~\cite{Song23Reaching, xing2023autonomous}. 
RL has demonstrated significant success in different domains of agile quadrotor control, ranging from time-optimal drone racing to obstacle avoidance~\cite{Song23Reaching, xing2024contrastive, kaufmann2022benchmark}.

State-of-the-art RL approaches have demonstrated specialization with great performance, even reaching the human-champion
level in single-task scenarios such as drone racing~\cite{kaufmann23champion}.
However, these specialized policies often struggle to perform novel, out-of-distribution tasks and require retraining from scratch when faced with even minor changes in task configuration~\cite{wang2024environment}. 
Consequently, the commonly used task-specific training approach is unsuitable for multi-task scenarios, making developing a multi-task RL policy for quadrotors a significant challenge.
A natural solution to MTRL problems is to train a network jointly on all tasks to uncover shared structures that can improve efficiency and performance beyond what individual task solutions can achieve.
However, learning multiple tasks simultaneously often poses a challenging optimization problem involving conflicting objectives~\cite{liu2021conflict}, sometimes resulting in poorer overall performance and data efficiency compared to individual task learning.

Despite these challenges, previous work on MTRL has demonstrated the potential of integrating shared task structures to perform various manipulation skills, such as lifting and pick-and-place, particularly in fixed-base manipulation scenarios~\cite{kalashnikov2022scaling, gupta2021reset, d2020sharing, aljalbout2024limt}.
While MTRL has shown promise in manipulation on a simulation benchmark~\cite{yu2020meta}, its application for quadrotors remains largely unexplored.
\vspace{-1em}
\subsection*{Contributions}
In this paper, we propose the first MTRL framework for learning various quadrotor control tasks efficiently.
A key advantage of MTRL is its ability to share knowledge across different tasks, thereby enhancing learning efficiency. 
Although the reward objectives of RL may vary between tasks, the underlying physical dynamics of the quadrotor, such as mass, inertia, and other physical properties, remain constant throughout the learning process. 
Leveraging these invariant conditions, our framework efficiently utilizes information sharing to learn multiple tasks within a single policy.

The reinforcement learning framework employs a multi-critic setup, and we propose a shared task encoder for observations containing dynamical information while handling task-specific observations using task-specific encoding networks.
By sharing common information and isolating task-specific data, we have developed a high-performance policy capable of executing maneuvers such as stabilization, velocity tracking, and racing.

We validate our MTRL policy in both simulation and real-world settings.
We demonstrate that our framework outperforms baseline approaches in terms of sample efficiency and overall task performance.
We believe this advancement is an important step toward developing a generalist quadrotor control policy, which could enable robots to handle diverse tasks in real-world scenarios more effectively.

%% file: figures/eyecatcher_vertical.tex
\begin{figure}[t!]
\centering
\vspace{2mm}
  \begin{tikzpicture}    
    \node [inner sep=0pt, outer sep=0pt] (img1) at (0.0,0) 
    {\includegraphics[width=0.9\linewidth]{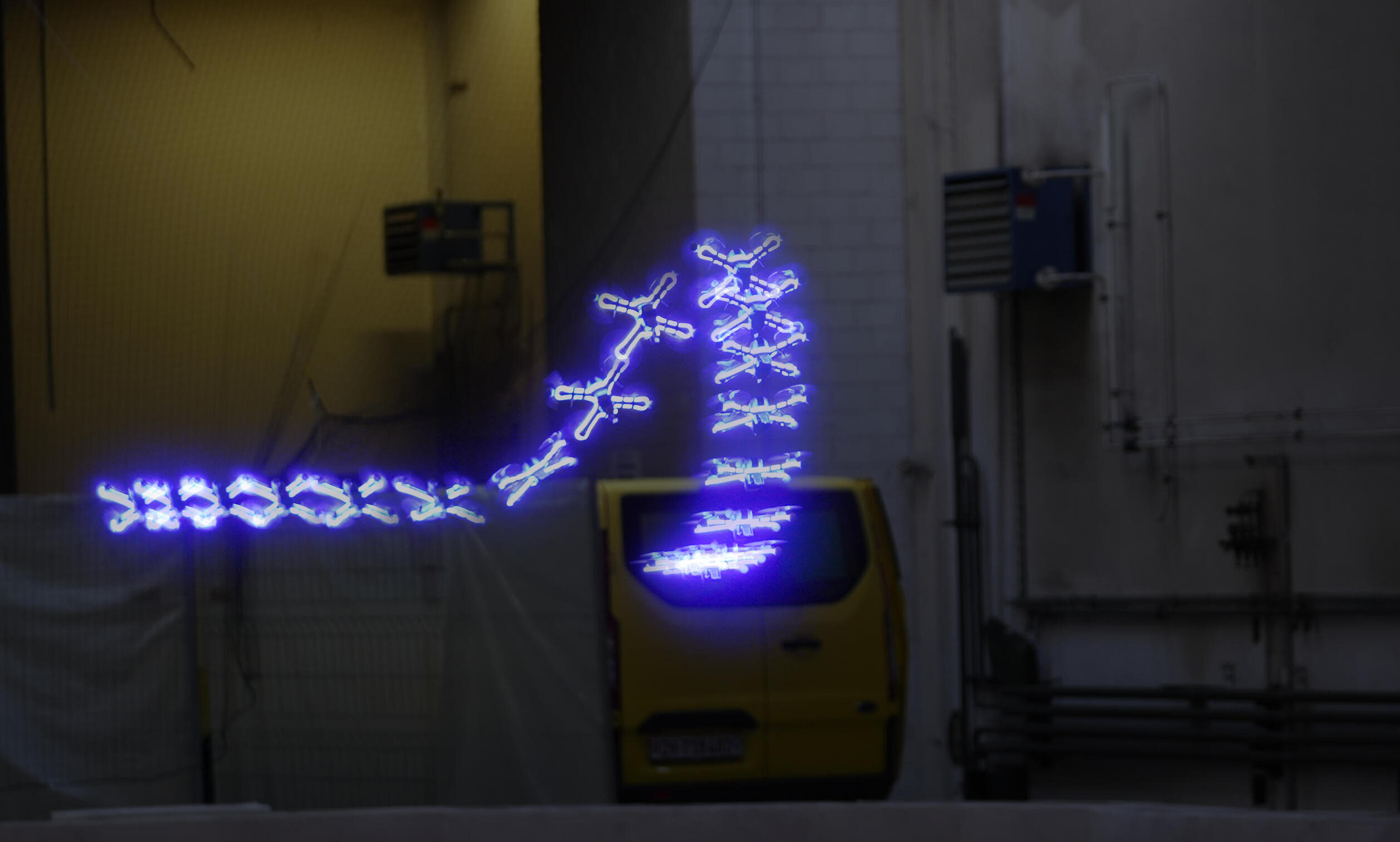}};
    \node [inner sep=0pt, outer sep=0pt, below=0.5mm of img1] (img2)
    {\includegraphics[width=0.9\linewidth]{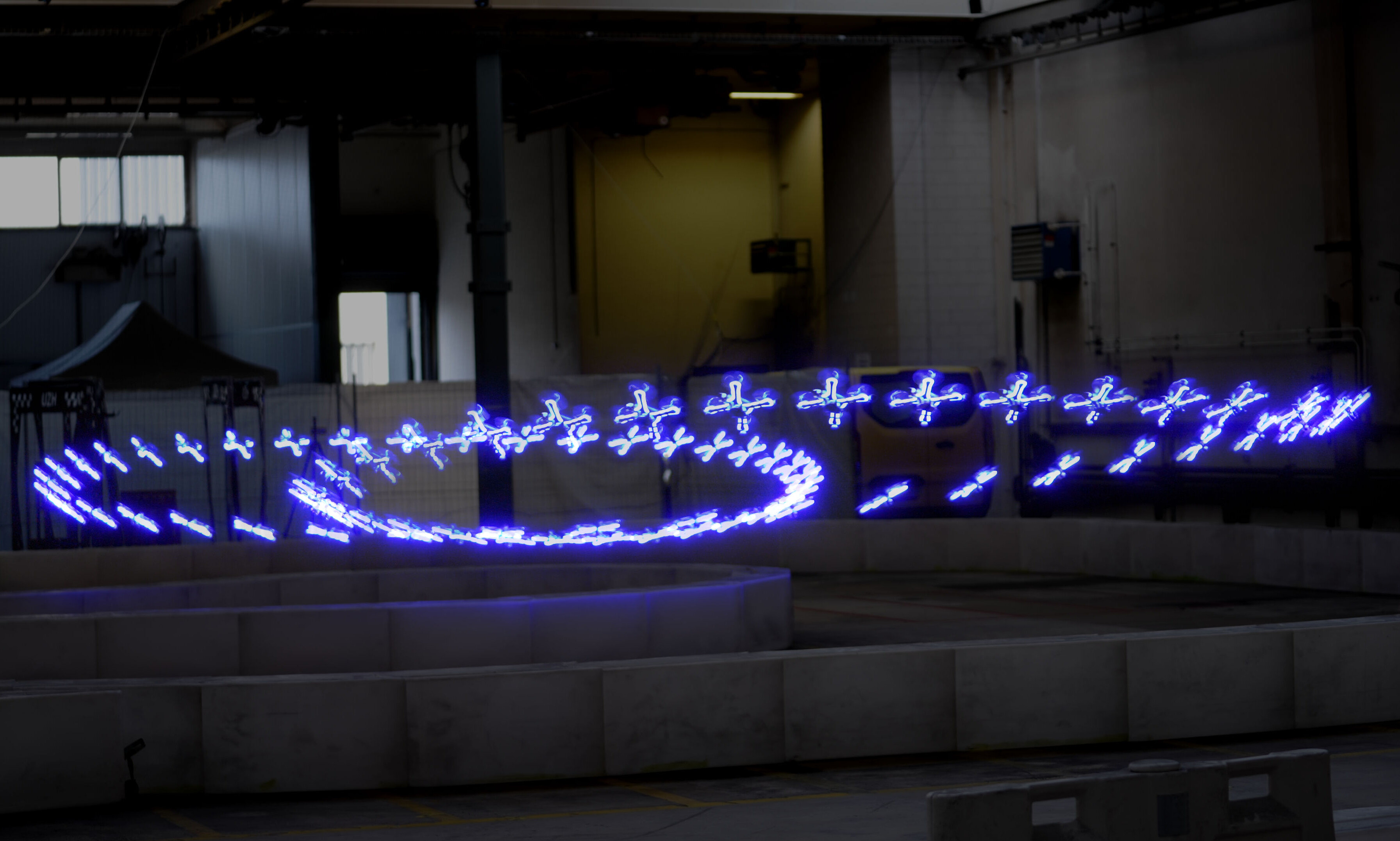}};
    \node [inner sep=0pt, outer sep=0pt, below=0.5mm of img2] (img3)
    {\includegraphics[width=0.9\linewidth]{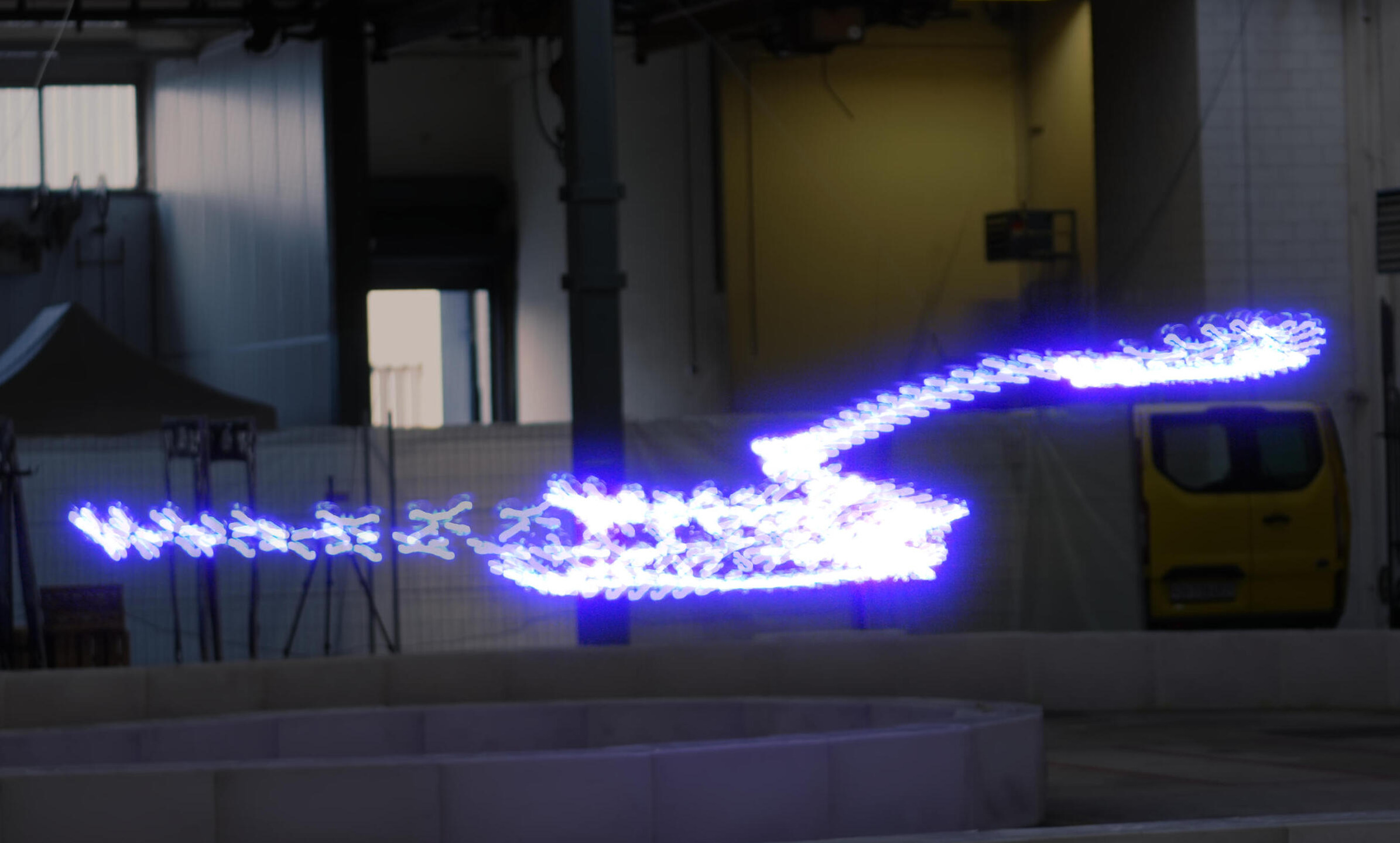}};
\end{tikzpicture}
	\caption{The proposed approach performs three distinct tasks for quadrotor control in the real world. The resulting \textit{single MTRL policy} can (\textit{Top})  stabilize the quadrotor from high speed, (\textit{Middle})  autonomously race through a fixed track, and (\textit{Bottom}) track randomly generated velocities.}
\label{fig:catcheye}
\vspace{-2em}
\end{figure}

%% file: sections/related_work.tex
\section{Related Works}
\subsection{Reinforcement Learning for Quadrotor Control}

In recent years, Reinforcement Learning (RL) has gained significant attention for the control of quadrotors. 
\ral{The work~\cite{hwangbo2017control} is one of the} first successful applications of RL to quadrotor control by tracking waypoints and recovering from challenging initial conditions. 
In~\cite{lambert2019low}, the performance gains of using RL for low-level control over classical methods are demonstrated. 
Besides obstacle avoidance~\cite{zhao2024learning, huang2024collision}, drone racing represents an important benchmark task for agile quadrotor flight, where the impact of reinforcement learning is very significant~\cite{Song23Reaching, kaufmann23champion}, even outperforming state-of-the-art model-based control~\cite{Song23Reaching}.
\ral{By optimizing several aspects of the training framework, in \cite{eschmann2024learning} it is demonstrated RL policies may be trained within seconds.
}
Several works focus on vision-based flight with reinforcement learning, such as~\cite{sadeghi2016cad2rl}, which uses a CAD model to produce discrete velocity commands directly from pictures. In ~\cite{xing2024bootstrapping},  it is shown how RL bootstrapped imitation learning~\cite{xing2024contrastive} benefits vision-based agile flight.
In ~\cite{kaufmann23champion}, the combination of vision-based state estimation and RL-based control enabled surpassing human world champions in drone racing.
In~\cite{geles2024demonstrating}, RL is used to learn drone racing from pixels without explicit state estimation.
\vspace{-1em}
\subsection{Multi-task Reinforcement Learning for Robotics}
\ral{
In~\cite{kalashnikov2022scaling}, a large-scale collective robotic learning system, can rapidly acquire diverse skills by sharing exploration, experience, and representations across tasks, improving overall performance and capabilities.
The work~\cite{gupta2021reset} demonstrates using MTRL to perform complex robotic skills like in-hand manipulation autonomously, significantly reducing the need for human resets during training sessions.
In~\cite{yang2020multi}, modularization in network design is explored to facilitate MTRL, improving both sample efficiency and the performance of robotic tasks by dynamically configuring network modules according to the task requirements.
In~\cite{xie2022lifelong}, a sequential multi-task learning scenario where a robot incrementally learns various tasks is demonstrated, using experiences from previous tasks to reduce the need for relearning and improve efficiency. 
Recently multi-task world models in proposed in\cite{aljalbout2024limt}, leveraging language model embeddings as task representations for model-based reinforcement learning of multiple robotic tasks.
However, most existing MTRL research focuses on manipulation tasks with varying underlying dynamics across tasks, limiting opportunities for knowledge sharing. 
In contrast, our work leverages the consistent dynamics of the quadrotor, enabling us to propose a novel framework that enhances knowledge sharing for more efficient multi-task learning.}

%% file: sections/methodology.tex
\section{Methodology}
\subsection{Notation}

In this manuscript, we define two reference frames. The first \( W \) is the fixed world frame with its \( z \)-axis aligned with gravity.
The second frame \( B \) is the quadrotor body frame. 
These reference frames are illustrated in Fig.~\ref{fig:reference_frames}. 
Vectors and matrices are represented as bold quantities, with capital letters denoting matrices. 
Vectors include a suffix indicating the frame in which they are expressed and their endpoint. 
For example, the quantity \(\bm{p}_{WB}\) represents the position of the body frame \( B \) relative to the world frame \( W \).
The rotation matrix that transforms a vector from frame \( B \) to \( W \) is denoted by \(\bm{R}_{WB}\). 

\subsection{Quadrotor Dynamics}
\input{figures/quadrotor}

Let \(\bm{p}_{WB}\), \(\bm{q}_{WB}\), and \(\bm{v}_{WB}\) represent the position, orientation, and linear velocity of the quadrotor, respectively, expressed in the world frame \(W\). 
Let \(\bm{\omega}_{B}\) denote the angular velocity of the quadrotor expressed in the body frame \(B\). 
Additionally, let \(c = \Sigma_i c_i\) represent the body’s collective thrust, where \(c_i\) is the thrust produced by the \(i\)-th motor, and let \(\bm{c} = \begin{bmatrix} 0 & 0 & c\end{bmatrix}^\intercal\) denote the collective thrust vector.
Here, \(m\) represents the mass of the quadrotor, and \(\bm{g}_W\) is the gravity vector.
Finally, let \(\bm{J}\) be the diagonal moment of inertia matrix, and \(\bm{\tau}_B\) the body’s collective torque. The quadrotor’s dynamic model is

\input{equations/dynamics}




\subsection{Policy Learning}
\subsubsection{Problem Formulation}
In the MTRL setting, we have $N$ tasks $\mathcal{T} = \{T_1, \cdots, T_i, \cdots, T_N\}$, where each task $T_i$ has a specific reward function $r_{i}$. 
Each task is defined by a Markov Decision Process (MDP) $\mathcal{M}_{i} = \left(\mathcal{S}_{i}, \mathcal{A}_{i}, \mathcal{P}_{i}, \mathcal{R}_{i}, \gamma_{i}\right)$, where $\mathcal{S}_{i}$ is the state space, $\mathcal{A}_{i}$ is the action space, $\mathcal{P}_{i}$ is the transition probability, $\mathcal{R}_{i}$ is the reward function, and $\gamma_{i}$ is the discount factor.
In the MTRL setting, the goal is to learn a policy $\pi$ that maximizes the expected return $J(\pi) $ across all tasks. 
Given uniformly sampled tasks, the optimization objective is defined as
\begin{equation}
    J(\pi) = \frac{1}{N} \sum_{i=1}^{N} \mathbb{E}_{\pi} \left[ \sum_{t=0}^{\infty} \gamma_{i}^t r_{i}(s_t, a_t) \right].
\end{equation}
\ral{
In our work, our multi-task configurations are primarily distinguished by the \emph{reward settings} $\mathcal{R}$ as model dynamics $\mathcal{P}$ remain identical due to the unchanging physical properties of the drones.
}
In the following sections, we detail the tasks considered in this work and the policy learning approach.

\subsubsection{Autonomous Racing}
The autonomous racing task can be formulated as an optimization problem that aims to minimize the time required for an agile quadrotor to navigate through a predefined sequence of gates~\cite{song2021autonomous}, as shown in Fig.~\ref{fig:racing}.
In this task, we use the observations $\bm{o} = \left[\bm{p}, \tilde{\bm{R}},  \bm{v}, \bm{\omega}, a_\text{prev}, 
\delta\bm{p}_1, \delta\bm{p}_2\right]$, where $\bm{p} \in \mathbb{R}^3$ denotes the drone's position, 
$\tilde{\bm{R}} \in \mathbb{R}^6$ is a vector comprising the first two columns of $\bm{R}_{\wfr\bfr}$~\cite{zhou2019continuity},
$\bm{v} \in \mathbb{R}^3$ and $\bm{\omega} \in \mathbb{R}^3$ denote the linear and angular velocity of the drone, 
$a_\text{prev}$ represents the previous action from the actor policy,
\ral{$\delta\bm{p}_1 \in \mathbb{R}^{12}$ represents the relative position differences of the four upcoming gate corners on the race track with respect to the drone agent, with each corner specified by a 3D position in the world frame.
Similarly, $\delta\bm{p}_2 \in \mathbb{R}^{12}$ represents the relative position differences of the corners from the next gate to the gate after that.
}
Here $\delta\bm{p}_1$ represents the difference of the 4 corners of the next gate to pass between the current quadrotor position. 
And $\delta\bm{p}_2$ represents the positional difference of the corners between the next gate to pass and the gate after the next gate to pass on the race track.
The RL policy training rewards are adjusted based on~\cite{Song23Reaching}.
The reward at time $t$, denoted as $r_t$, is defined as the sum of various components
 \begin{equation}
	 r_t^{\mathrm{racing}} = r_t^{\mathrm{prog}} + r_t^{\mathrm{perc}} + r_t^{\mathrm{act}} + r_t^{\mathrm{br}} + r_t^{\mathrm{pass}} + r_t^{\mathrm{crash}},
 \end{equation}
where $r_t^{\mathrm{prog}}$ encourages progress towards the next gate to be passed~\cite{Song23Reaching},
$r_t^{\mathrm{perc}}$ encodes perception awareness by adjusting the quadrotor’s attitude such that the optical axis of its camera points towards the next gate’s center, $r_t^{\mathrm{act}}$ penalizes action changes from the last time step, 
$r_t^{\mathrm{br}}$ penalizes body rates and consequently reduces motion blur, 
$r_t^{\mathrm{pass}}$ is a binary reward that is active when the robot successfully passes the next gate,
$r_t^{\mathrm{crash}}$ is a binary penalty that is only active when a collision happens, which also ends the episode. 
The reward components are formulated as follows
\begin{equation}
    \begin{aligned}
        r_t^{\mathrm{prog}} & = \alpha_1(d_{\mathrm{Gate}}(t-1) - d_{\mathrm{Gate}}(t)), \\
        r_t^{\mathrm{perc}} & = \alpha_2 \exp(\alpha_3 \cdot \delta_\mathrm{cam}^4), \\
        r_t^{\mathrm{act}} & = \alpha_4 \lVert\bm{u}_t - \bm{u}_{t-1}\rVert, \\
        r_t^{\mathrm{br}} & = \alpha_5 \lVert\bm{\omega}_{\mathcal{B},t}\rVert, \\
        r_t^{\mathrm{pass}} & =  \alpha_6  \quad\text{if robot passes the next gate}, \\
        r_t^{\mathrm{crash}} & = \alpha_7 \quad\text{if robot crashes (gates, ground)}. 
    \end{aligned}
    \vspace{-0.4em}
\end{equation}

\input{figures/overview}
\subsubsection{Stabilization from High Speed}
In the stabilization task, the quadrotor is expected to recover to the static status, given randomized poses and high initial velocities. 
\ral{Here, successful stabilization is defined as achieving near-zero velocities around a predefined height $z_d$, starting from random initial positions and velocities.}
In this setting, the quadrotor is initialized with a random position, orientation, and linear and angular velocities. The observation contains $\bm{o}_h =[\bm{p}, \tilde{\bm{R}}, \bm{v}, \bm{\omega}, a_\text{prev}, \ddot{\bm{p}}_{WB}, z_d]$.
\ral{Here $ \ddot{\bm{p}}_{WB}$ is defined as the acceleration of the quadrotor agent in the world frame, and $z_d$ represents a predefined constant height in meters to stabilize at.}
The reward function $r_t^{\mathrm{stabilize}}$ is defined as
\begin{equation}
	r_t^{\mathrm{stabilize}} = r_t^{\mathrm{height}} + r_t^{\mathrm{attitude}} + r_t^{\mathrm{velocity}} + r_t^{\mathrm{br}} + r_t^{\mathrm{act}} + r_t^{\mathrm{success}}.
\end{equation}
Here $r_t^{\mathrm{height}}$ rewards the quadrotor for maintaining a constant height, $r_t^{\mathrm{attitude}}$ rewards the quadrotor for maintaining a constant orientation, $r_t^{\mathrm{velocity}}$ and $r_t^{\mathrm{angular}}$ penalizes the non-zero linear and angular velocities, $r_t^{\mathrm{act}}$ penalizes the non-smooth actions, and $r_t^{\mathrm{success}}$ is a discrete reward when the quadrotor is stabilized. 
All the rewards are formulated here using the $\mathcal{L}_2$ norm multiplied by a corresponding constant coefficient.
The detailed rewards are defined as
\begin{equation}
    \begin{aligned}
        r_t^{\mathrm{height}} & = \beta_1 \lVert z_t - z_d\rVert, \\
        r_t^{\mathrm{attitude}} & = \beta_2 \lVert\bm{R}_t\rVert, \\
        r_t^{\mathrm{velocity}} & = \beta_3 \lVert\bm{v}_t\rVert, \\
        r_t^{\mathrm{br}} & = \beta_4 \lVert\bm{\omega}_{\mathcal{B},t}\rVert, \\
        r_t^{\mathrm{act}} & = \beta_5 \lVert\bm{u}_t - \bm{u}_{t-1}\rVert, \\
        r_t^{\mathrm{success}} &= \beta_6\quad\text{if robot remains hovering}.
    \end{aligned}
\end{equation}
During training, we applied a curriculum to gradually increase the difficulty of the tasks.
We increase the initial speed of the quadrotor in the \(x\), \(y\), and \(z\) axes by 10\% for every 100,000 data samples. The curriculum will stop once the predefined upper limits in each direction are reached.

\subsubsection{Velocity Tracking}
In the velocity tracking task, the quadrotor is required to track a randomly generated velocity profile. 
The observation contains $\bm{o}_v = [\bm{p}, \tilde{\bm{R}}, \bm{v}, \bm{\omega}, a_\text{prev}, \bm{v}_d, \ddot{\bm{p}}_{WB}]$, where $\bm{v}_d$ represent the desired linear velocity.
The reward function $r_t^{\mathrm{tracking}}$ is defined as
\begin{equation}
	r_t^{\mathrm{tracking}} = r_t^{\mathrm{velocity}} + r_t^{\mathrm{br}} + r_t^{\mathrm{act}},
\end{equation}
where $r_t^{\mathrm{velocity}}$ rewards the quadrotor for tracking the desired velocity,
$r_t^{\mathrm{br}}$ penalizes the quadrotor for non-zero angular velocities, and $r_t^{\mathrm{act}}$ penalizes the quadrotor for excessive actions.
All the rewards are formulated here using the $\mathcal{L}_2$ norm multiplied by a corresponding constant coefficient,
\begin{equation}
    \begin{aligned}
        r_t^{\mathrm{velocity}} & = \lambda_1 \lVert\bm{v}_t - \bm{v}_{d}(t)\rVert, \\
        r_t^{\mathrm{br}} & = \lambda_2 \lVert\bm{\omega}_{\mathcal{B},t}\rVert, \\
        r_t^{\mathrm{act}} & = \lambda_3 \lVert\bm{u}_t - \bm{u}_{t-1}\rVert. \\
    \end{aligned}
\end{equation}
During training, we implemented a curriculum to progressively increase the difficulty of the tasks. 
The desired speed of the quadrotor in the 
\(x\), \(y\), and \(z\) axes was increased by 1\si{\meter\per\second} for every 100,000 data samples.
This gradual increase will continue until the predefined upper-speed limits in each direction are achieved.

\input{figures/rewards}
\vspace{-1em}
\subsection{Multi-task Learning Framework}
Since the control or navigation tasks are typically not contact-rich for a fixed quadrotor platform, we generally do not expect changes in the dynamics equations (see Eq.~\ref{eq:dynamics}), whether during different phases of a single task or across various tasks.
Many previous works also present this valid assumption, from agile drone racing to obstacle avoidance~\cite{Song23Reaching, kaufmann2022benchmark, kaufmann23champion}.
Hence, in the quadrotor setting, we can utilize the consistent physical property to simplify the multi-task scenario by assuming an identical transition probability across all tasks.
This assumption serves as the primary motivation for our proposed information-sharing structure in the MTRL setup, allowing us to share data samples across tasks to efficiently learn an encoding network for the observation information related to the transition dynamics.
To leverage the shared information across tasks, we propose a multi-task learning framework that consists of shared and task-specific modules. 

As we aim to have a single policy capable of solving multiple tasks, our actor needs to be common to all tasks.
As shown in the previous sections, the observation space in different tasks shares some common features, namely the position, orientation, linear and angular velocities, and the previous action.
The selection of the shared features is based on the fact that these represent the essential dynamical properties outlined in Eq.~\ref{eq:dynamics}
, which remain consistent across different tasks.
Other features can be task-specific and have different dimensions, which would lead to different tasks having different policy input sizes.
To distinguish observations from various potential tasks, we use an identifier based on the task-specific observation length, which allows us to incorporate one-hot encoding into the task-specific observations.
To overcome this, we propose a feature encoding architecture including a shared encoder network among different tasks to extract the shared features from the observation. Meanwhile, we use task-specific encoders to generate the task-specific policy inputs.
Although the task-specific observations' dimensions vary from task to task, the task-specific encoder is designed to map the specific observation to the same latent space.

For the MTRL part, we employ a shared actor policy with multiple individual critic networks, each corresponding to a specific task.
The actor policy takes the concatenated latent features from the shared encoder and the task-specific encoder as input and outputs the action, namely the collective thrust and the body rates~\cite{kaufmann2022benchmark}.
The overall architecture is shown in Fig.~\ref{fig:overview}.
We train the shared actor policy to maximize the expected return across all tasks, while the individual critic networks are trained to evaluate the value function for each task.
In contrast, the task-specific critic networks take directly the full observation as the input and output of the value function for each task. 
Based on ~\cite{andrychowicz2020matters}, we went for the choice not sharing the policy feature encoder with the critic networks, as it has been shown to improve the performance of the policy.

%% file: figures/quadrotor.tex
\begin{figure}[t!]
    \centering
    \includegraphics{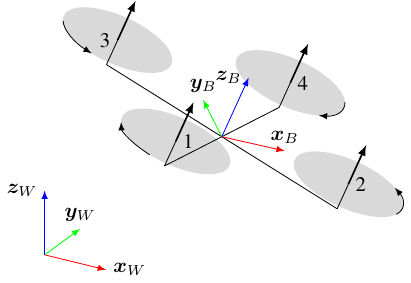}
    \caption{Diagram of quadrotor model \ral{with the world and
body frames and propeller numbering convention.}}
    \label{fig:reference_frames}
    \vspace{-1.5em}
\end{figure}

%% file: equations/dynamics.tex
\begin{equation}
	\begin{aligned}
		\dot{\bm{x}} = \begin{bmatrix}
			\dot{\bm{p}}_{WB} \\
			\dot{\bm{q}}_{WB} \\
			\dot{\bm{v}}_{WB} \\
			\dot{\bm{\omega}}_{B} \\
		\end{bmatrix} = \begin{bmatrix}
		\bm{v}_{WB}\\
		\frac{1}{2} \Lambda({\omega}_B) \cdot \bm{q}_{WB}\\
		\bm{q}_{WB} \odot \bm{c}/m + \bm{g}_W\\
		\bm{J}^{-1}(\bm{\tau}_B - \bm{\omega}_B\times \bm{J} \cdot \bm{\omega}_B)
	\end{bmatrix}.
	\end{aligned}
 \label{eq:dynamics}
\end{equation}

%% file: figures/overview.tex
\begin{figure*}[!htp]
	\centering
	\includegraphics{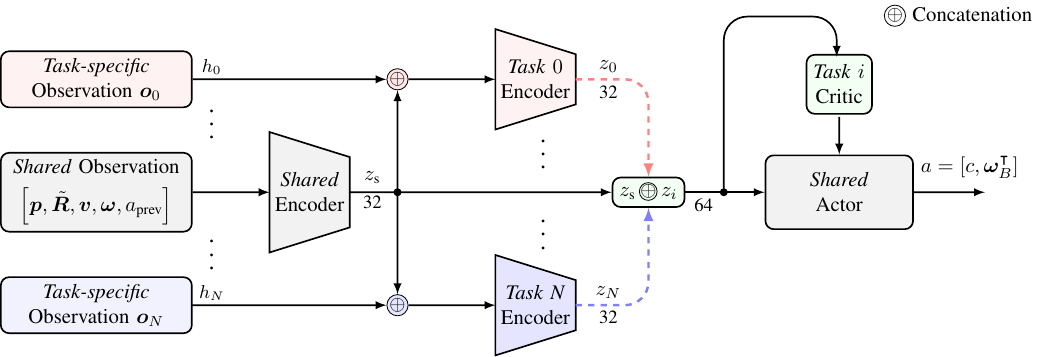}
	\caption{
 Our MTRL framework utilizes a shared encoder for observations related to the quadrotor dynamics across all tasks. The embedding output from the shared encoder is then merged with the task-specific observation (e.g., the gate observation from the racing task and the desired velocity from the tracking task) to create a task-specific embedding. 
 The policy uses both the concatenated embedding (64) from the shared embedding (32) and the task-specific embedding (32) to generate control commands. A separate critic function is used for each task, which is not employed during deployment.}
	\label{fig:overview}
 \vspace{-0.3em}
\end{figure*}

%% file: figures/rewards.tex
\begin{figure*}[t!]
    \centering
    \includegraphics[width=\linewidth]{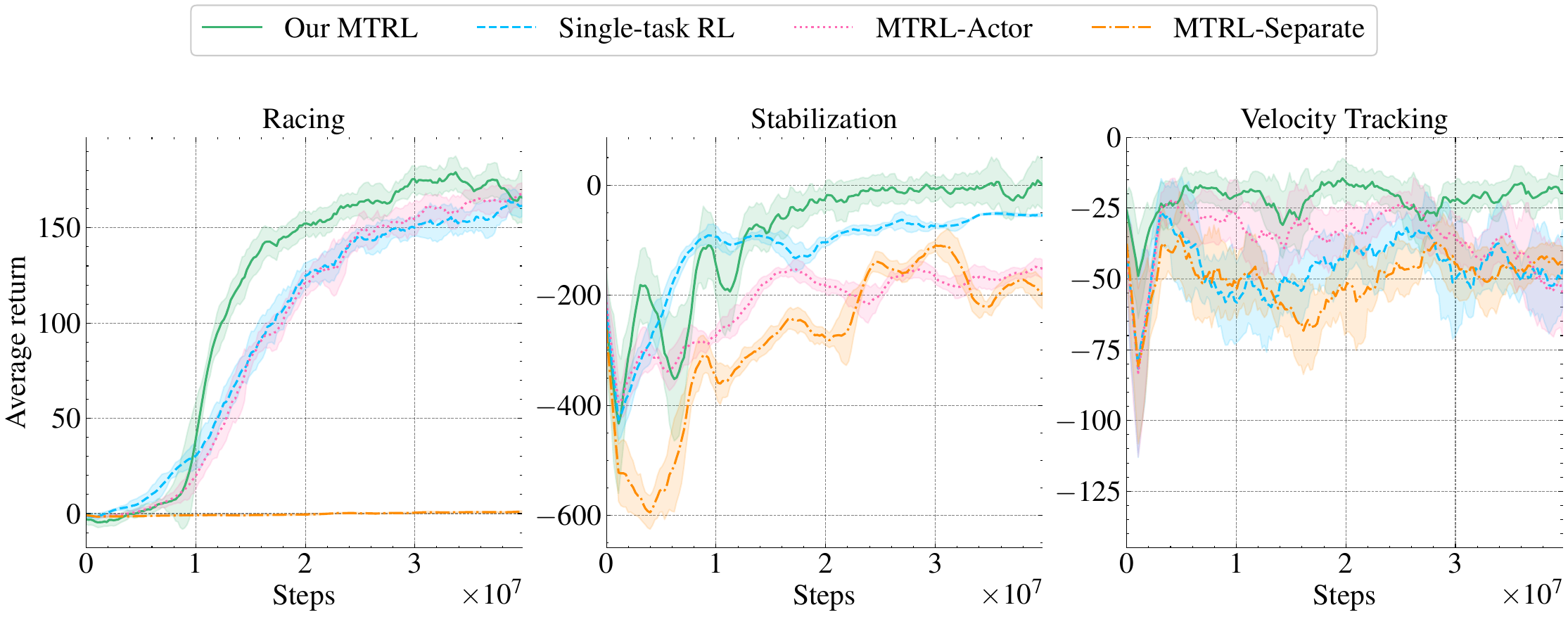}
    \caption{Overview of the average return comparison of different tasks. \ral{It is clearly shown that our proposed MTRL approach achieves a higher average return within the same number of training steps compared to single-task RL baselines. Notably, single-task RL policies still perform comparably to the MTRL approach when only the actor network is shared.}}
    \label{fig:rewards_comparison}
    \vspace{-1.5ex}
\end{figure*}

%% file: sections/experiments.tex
\section{Experiments}
Using the individual tasks described in the previous section, we evaluate the performance of the proposed MTRL approach. 
Our experiments are designed to answer the following research questions: 
(i) How sample-efficient is our MTRL approach compared to the single-task RL baselines? (ii) How does the MTRL policy's performance compare to the single-task RL policies? 
(iii) How do different knowledge-sharing strategies affect the MTRL performance? 
(iv) Does the result transfer to a real-world scenario?
\vspace{-1em}
\subsection{Training Configurations}
For the policy training, we employ a policy network consisting of a two-layer MLP, each layer containing 256 neurons, with a final layer outputting a 4-dimensional vector using a $tanh$ activation function.
In our experiments, for the shared dynamic encoder, we use a three-layer MLP with 19 neurons in the input layers and 128 neurons in the hidden layers to generate a 32-dimensional latent embedding.
For the task-specific encoder, we use a three-layer MLP with task-dependent input dimensions and 128 neurons in the hidden layer to generate a 32-dimensional latent embedding.
Table~\ref{tab:params} shows the detailed task reward parameters for the MTRL training. 
In our setting, we optimized our hyperparameters solely based on the single-task performance.
For MTRL training, we employ model-free reinforcement learning
approach using \ral{Proximal Policy Optimization (PPO)}~\cite{schulman2017proximal}.
For the quadrotor platform used for both training and deployment, we detail the information regarding components and physical parameters in Table \ref{tab:physical_drone}. We use the Flightmare simulator~\cite{yunlong2020flightmare} for policy training in simulation. 
\input{tables/hyperparam}

\input{tables/quadrotor.tex}

\input{tables/performance}

\vspace{-1em}
\subsection{Baselines}
In our experiments, we compare our approach with the following baselines:
(i) \textbf{Single-task RL}: We train a separate policy for each task using the same RL algorithm as the MTRL approach.
(ii) \textbf{MTRL-Actor}: For this baseline, we keep the \textit{Actor} as the only network shared among different tasks.
And all the other encoder networks and critic networks are different among all tasks.
We train a shared actor policy with multiple individual critic networks, each corresponding to a specific task. 
(iii) \textbf{MTRL-Seperate}: To ablate the design choice of fusing shared and task-specific observations, we train a shared encoder and task-specific encoders, but the task encoder does not receive the shared observation as input.
The observation encoders are different for each task, and the shared actor policy takes the latent embedding directly as input to output the action.
\ral{For all of the approaches, we performed 10 runs of the same training configurations using different random seeds, and we report the average evaluation metrics in this section.}
Since the primary contribution of our approach is to enhance learning efficiency and task performance of the RL policies using MTRL, we have chosen not to include baselines that rely solely on classical control approaches in our comparison.
\vspace{-2em}
\subsection{Sample Efficiency Analysis}
To evaluate the sample efficiency of our MTRL approach, we compare the return curves of the MTRL approach with the aforementioned baselines.
All the policies are trained using the same hyperparameters and the same number of training steps.
The number of training steps in our setting is 40M, where all the policies' performances converge.
As shown in Fig.~\ref{fig:rewards_comparison}, the MTRL approach outperforms the single-task RL baselines in all three different tasks.
The MTRL approach achieves a higher average return in the same number of training steps than single-task RL baselines.
Notably, the single-task RL policies still perform closely to the MTRL approach when only sharing the actor network.
This highlights the necessity of information sharing in our framework.
However, although information sharing is beneficial, the MTRL-Seperate cannot fly at all in the racing task. 
This is likely because if we do not fuse the shared information with task-specific information, it becomes difficult when the task requirements are conflicting, e.g. racing and stabilization from high speed. 
The policy will then prefer learning the rather easier task.
Hence, this highlights the importance of fusing the shared and task-specific information in the MTRL framework.

\input{figures/racing}

\subsection{Individual Task Performance of MTRL Policy}
In this section, we showcase the individual task performance of our MTRL policy.
We evaluate the performance of the MTRL policy in three different tasks: stabilization from high speed, racing, and velocity tracking.
The MTRL policy is trained using the same hyperparameters as the single-task RL policies.
\input{figures/tasks}

\subsubsection{Racing Performance}
In the racing task, we evaluate the performance of the MTRL policy by racing on a predefined race track. The race track contains 6 gates, and the quadrotor is required to pass these gates in a fixed order.
The policy is evaluated in 64 different starting positions in uniformly sampled starting positions. Fig.~\ref{fig:racing} visualizes one sample rollout that successfully completes the race track.

\subsubsection{Stabilization Performance}
As shown in Fig.~\ref{fig:gradcam} (a), we evaluate the high-speed stabilization task by randomly initializing the quadrotor with various positions, orientations, and velocities.
The initial speeds in the \(x\) and \(y\) directions are randomly sampled from \([-20, 20]\) \si{\meter\per\second}, while the velocity in the \(z\) direction is sampled from \([-4, 4]\) \si{\meter\per\second}.
To prevent the quadrotor from crashing into the ground when the initial \(z\)-direction velocity is high and negative, we adjust the initial height accordingly to ensure the quadrotor will not crash within one second, even if no control input is applied.
From the results, we observe that the MTRL policy can successfully stabilize the quadrotor in the hovering condition from high speed task within seconds, even when the initial conditions are challenging; the maximum initial speed from our experiments is 82.57 \si{\kilo\meter\per\hour}.
This demonstrates the robustness of the MTRL policy in stabilization from high-speed tasks.

\subsubsection{Velocity Tracking Performance}
In the tracking task, we evaluate the performance of the MTRL policy by tracking a randomly generated velocity command, where we apply random walks in the acceleration space.
\ral{We randomize the velocity references in $x$ and $y$ directions up to 54\si{\kilo\meter\per\hour} and 18\si{\kilo\meter\per\hour} in $z$ direction.
}
As shown in Fig.~\ref{fig:gradcam} (b), the MTRL policy can successfully track the challenging commanded velocities,  where they can go up to 50 \si{\kilo\meter\per\hour} in a very short time, and we did not even include the non-holonomic constraints of the quadrotor to generate velocity command.
\vspace{-1em}
\subsection{Quantitative Analysis}
\ral{
\subsubsection{Evaluation Metrics}
For the autonomous racing task, we use three evaluation metrics: success rate (SR), mean-gate-passing-error (MGE), and lap time (LT). 
SR is the ratio of completed laps to total trials. MGE measures the distance between the drone's position and the gate center when passing through; here, the inner gate size used for experiments is 1.5m.
LT indicates the duration of completing a full race track and flying through all gates. 
For the stabilization task, we employ two evaluation metrics, namely $t_\text{half}$ and $t_\text{full}$, to evaluate the time usage of the RL controller to stabilize the quadrotor. 
Here $t_\text{half}$ is the time usage to reduce the actual velocity to half of the initial velocity, and $t_\text{full}$ is the time usage to control a quadrotor to hover condition; in our experiment, we determine this when the quadrotor's linear velocity is smaller than $0.5 \si{\meter\per\second}$.
For the velocity tracking task, we simply use a tracking error metric $\bm{e_v}$, which computes the averaged velocity difference over time.
}
\subsubsection{Analysis}
In Table~\ref{tab:performance}, we present a quantitative comparison of our approaches with the baselines.
To demonstrate the effectiveness of the policy trained with different numbers of samples, we list the performance of the MTRL policies in two different timesteps, namely 20M and 40M.
First of all, for the policy trained with 20M steps, our MTRL approach demonstrates a much better task performance than all of the baseline approaches. 
This strongly demonstrates the sample efficiency of our approach.
Secondly, when the policy is trained till convergence, the policy's performance of our approach is still not worse than learning individually in all the tasks and all the metrics.
This further indicates the claim of our approach: our MTRL framework could learn multiple tasks efficiently without trading off performance.
\vspace{-0.4em}
\subsection{Real World Performance}
To demonstrate the effectiveness of our policy improvements, we conducted validation tests in real-world scenarios.
We utilized an Agilicious quadrotor platform~\cite{foehn2022agilicious} with the identical properties presented in~\ref{tab:params} with state estimation provided by a VICON motion capture system to feed accurate inputs to the policy. 
For low-level control, the BetaFlight2 firmware was employed to track the commanded collective thrusts and body rates.
We conducted five individual runs for each task, varying the starting conditions in each run.
Remarkably, our MTRL policy achieved a 100$\%$ success rate across all tasks in these real-world experiments, as illustrated in Fig.~\ref{fig:catcheye}.
These results clearly indicate that our policy effectively transfers to and performs reliably in real-world scenarios.
The detailed evaluation results are presented in Table \ref{tab:realworld}. 
The metrics for each task indicate that our policy consistently delivers stable performance both in simulation and the real world using identical configurations.
\input{tables/realworld}

%% file: tables/hyperparam.tex
\begin{table}[b!] 
   \centering 
      \caption{Reward parameters for MTRL training.} 
    \fontsize{8pt}{10pt}\selectfont
    \begin{tabular}{ cc|cc|cc }
        \toprule
          \textbf{Param.}  & \textbf{Value} &\textbf{Param.}  & \textbf{Value}&\textbf{Param.}  & \textbf{Value}  \\ 
          \midrule
          \grayrow
          $\alpha_1$   & 0.5& $\beta_1$   & -2e-3 & $\lambda_1$   & -2e-4  \\
           $\alpha_2$   & 0.025&  $\beta_2$   & -2e-4 & $\lambda_2$   & -1.2e-3  \\ 
           \grayrow
           $\alpha_3$   & -1.0&  $\beta_3$   & -4e-5 & $\lambda_3$   & -1e-4 \\
            $\alpha_4$   & -2e-4 & $\beta_4$   & -1e-5 &   &   \\
            \grayrow
             $\alpha_5$   & -5e-4& $\beta_5$   & -1e-4 &   &  \\
              $\alpha_6$   & -5 & $\beta_6$   & 10&   &  \\
              \grayrow
               $\alpha_7$   & -10 &  & &   &    \\
        \bottomrule
    \end{tabular}
   \vspace{0.5em} 
   \label{tab:params}
\end{table}

%% file: tables/quadrotor.tex
\begin{table}[t!] 
   \centering 
      \caption{Overview of the drone parameters for both simulation and real-world experiments.} 
    \fontsize{9pt}{12pt}\selectfont
    \begin{tabular}{ l  c }
        \toprule
          \textbf{Param.}  & \textbf{Agilicious Platform}  \\ \cline{1-2}
          \grayrow
          Thrust-to-weight Ratio   & 5.78  \\
          Mass [\SI{}{\kilogram}]   &    0.6\\
          \grayrow
          Maximum Thrust [\SI{}{\newton}]  & 20.00   \\
          Arm Length [\SI{}{\meter}]  & 0.15 \\
          \grayrow
          Inertia  [\SI{}{\gram\square\meter}]  & [2.50, 2.51, 4.32] \\
          Motor Time Constant [\SI{}{\second}]  & 0.033 \\ 
        \bottomrule
    \end{tabular}
   \vspace{-1.5em} 
   \label{tab:physical_drone}
\end{table}

%% file: tables/performance.tex
\begin{table*}[!ht]
    \centering
    \vspace*{2pt}
    \caption{Individual task performance after 20M and 40M training samples. For the stabilization task, we evaluate the time to reduce the velocity to half the initial velocity ($t_{\mathrm{half}}$) and to hovering ($t_{\mathrm{full}}$). For velocity tracking, we report the tracking error $\bm{e_v}$. 
    For racing, we report the success rate (SR), the mean gate passing error (MGE), and the lap time (LT).} 
    \setlength{\tabcolsep}{3.5pt}
    \begin{tabular}{c|cccccc|cccccc}
    \toprule
     \multirow{3}*{\textbf{Methods}} &\multicolumn{6}{c|}{\textbf{20M Samples}}&\multicolumn{6}{c}{\textbf{40M Samples}}\\
    &\multicolumn{3}{c}{\textit{Racing}}&\multicolumn{1}{c}{\textit{Stabilization}}&& \multicolumn{1}{c|}{\textit{Tracking}}  & \multicolumn{3}{c}{\textit{Racing}}& \multicolumn{1}{c}{\textit{Stabilization}}&& \multicolumn{1}{c}{\textit{Tracking}}\\
    &SR [\%] &MGE [\SI{}{\meter}]  &  LT [\SI{}{\second}] &$t_{\mathrm{half}}$ [\SI{}{\second}]&$t_{\mathrm{full}}$ [\SI{}{\second}]&$\bm{e_v}$  [\SI{}{\meter\per\second}] &SR [\%] &MGE [\SI{}{\meter}]  &  LT [\SI{}{\second}]&$t_{\mathrm{half}}$[\SI{}{\second}]&$t_{\mathrm{full}}$[\SI{}{\second}] &$\bm{e_v}$[\SI{}{\meter\per\second}]  \\
    
    \midrule
    \grayrow
    Single-task RL & 95& 0.224& 6.23 & 0.50 & 4.20 &2.15& \textbf{100}& 0.163& 5.83 & 0.41 & 3.53 &\textbf{1.18}\\
    MTRL-Actor & 94& 0.231& 6.09& 0.61 & 4.57 &2.22& \textbf{100}& 0.159& 5.88  & 0.39 & 3.29 &1.44\\
    \grayrow
    MTRL-Separate & 0& crash& crash& 0.59 & 4.29&2.04 & 0& crash& crash& 0.91 & 4.12 &1.93\\
    \textbf{Ours} & \textbf{100}& \textbf{0.187}& \textbf{5.92}& \textbf{0.41} & \textbf{3.77} &\textbf{1.69} & \textbf{100}& \textbf{0.152}&  \textbf{5.80}&\textbf{0.35} & \textbf{3.26} &1.21 \\
    \bottomrule
\end{tabular}
    \label{tab:performance}
\vspace*{-13pt}
\end{table*}

%% file: figures/racing.tex
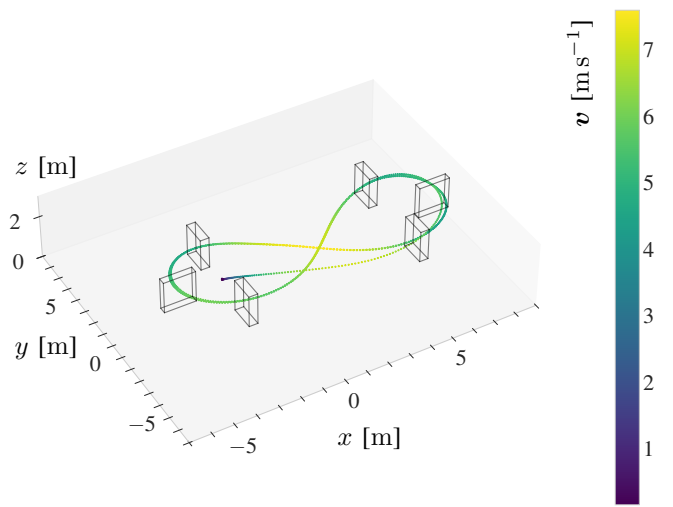
\begin{figure}[b!]
    \centering
     \vspace{-1em}
     \begin{tikzpicture}
\node [inner sep=0pt, outer sep=0pt] (img1) at (0,0) 
{\includegraphics[width=0.98\linewidth]{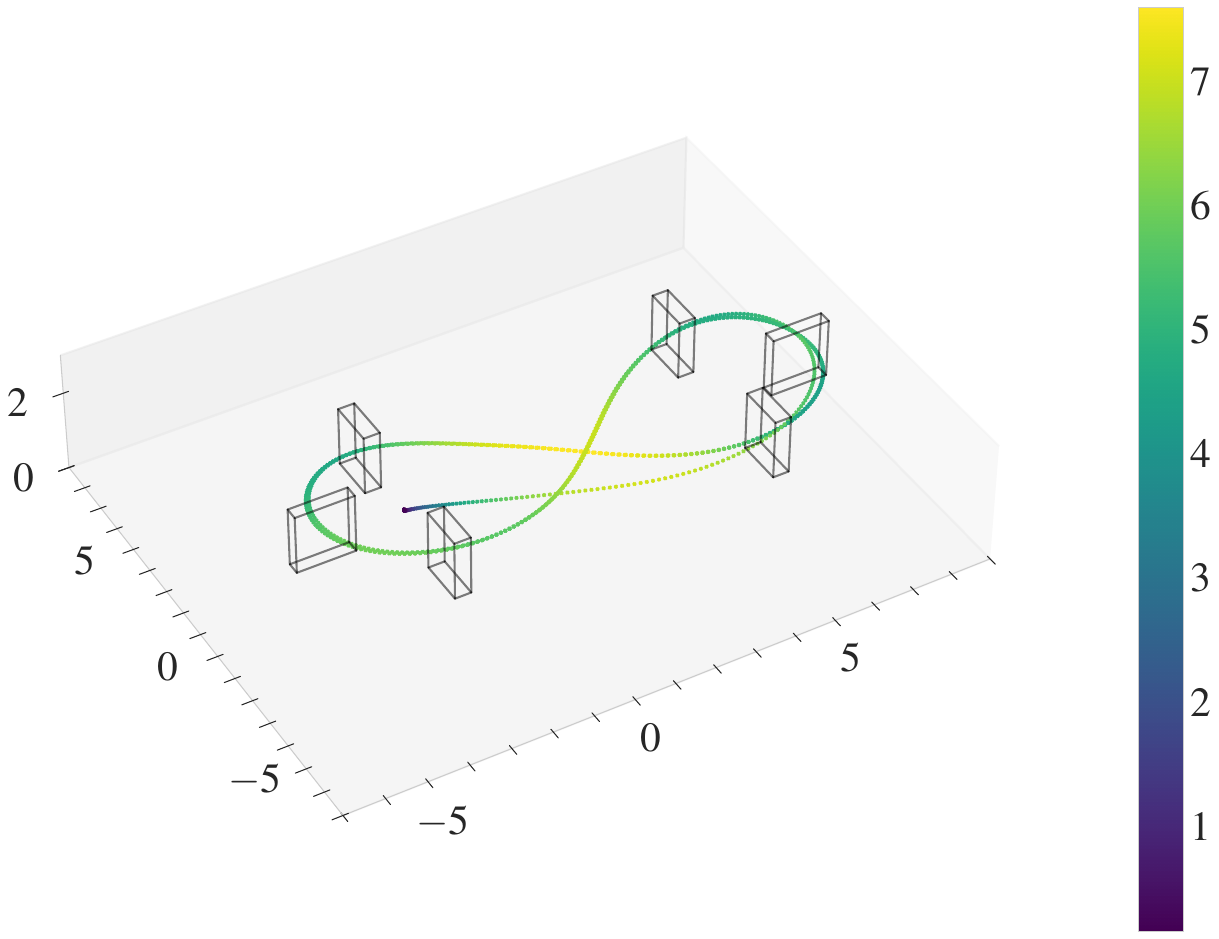}};
\node [rotate=0] at (0.5, -2.4) {$x$ [\si{\meter}]};
\node [rotate=0] at (-3.8, -1.2) {$y$ [\si{\meter}]};
\node [rotate=0] at (-3.8, 1.2) {$z$ [\si{\meter}]};
\node [rotate=90] at (3.3, 2.45) {$\bm{v}$ [\si{\meter\per\second}]};

    \end{tikzpicture}

    \caption{Illustration of one racing policy rollout. \ral{The policy successfully completes a Figure-8 race track, which consists of six gates, with a 100\% success rate.}}
    \label{fig:racing}
\end{figure}

%% file: figures/tasks.tex
\begin{figure*}[t!]
    \centering

\begin{tikzpicture}[>=stealth]
\node [inner sep=0pt, outer sep=0pt] (img1) at (0,0) 
{\includegraphics[width=0.45\linewidth]{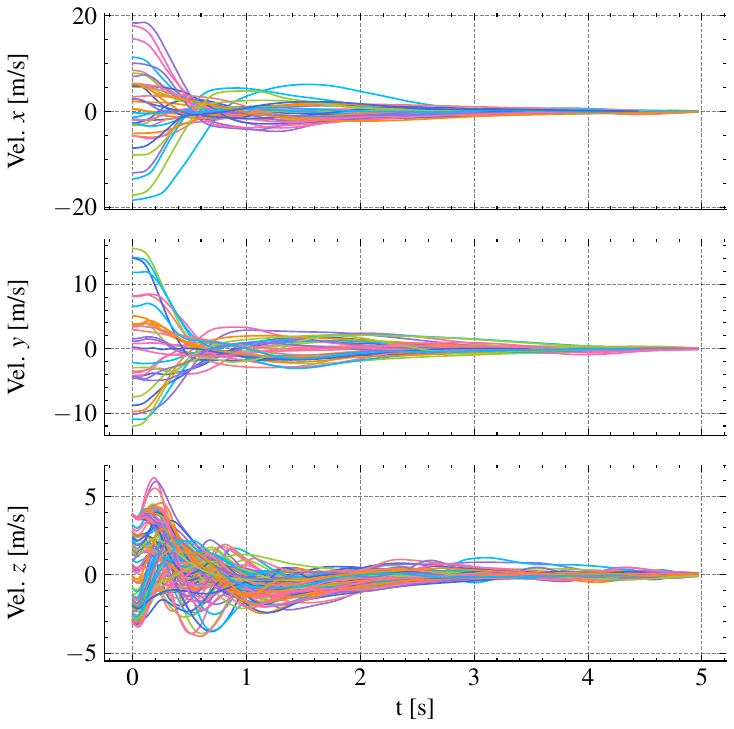}};
\node (img3)at (9, 0.46) {\includegraphics[width=0.45\linewidth]{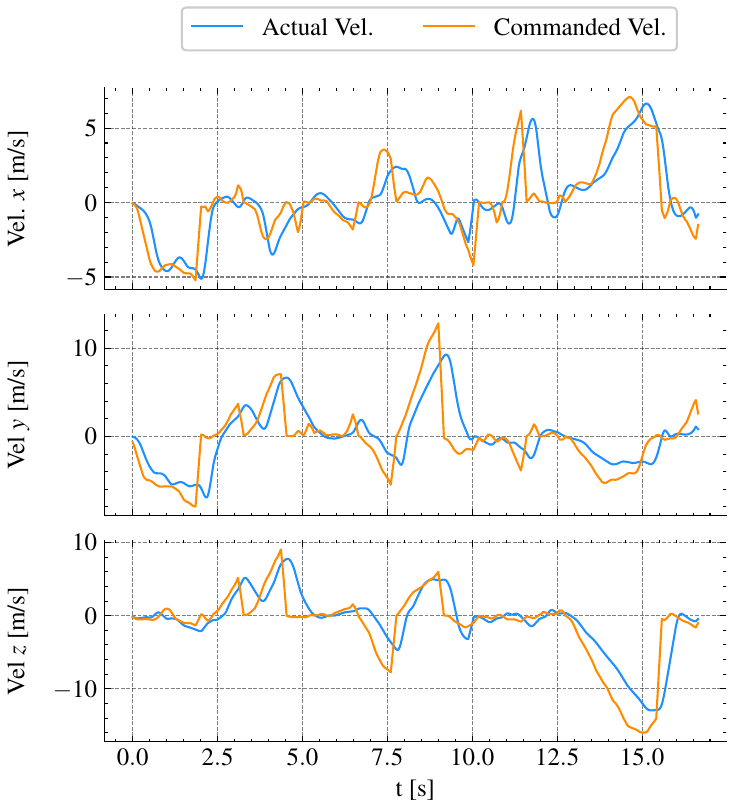}};

\node(text1) at (0.5, -4.5){(a) Stabilization Performance};
\node(text1) at (9.5, -4.5)(text1){(b) Tracking Performance};
\end{tikzpicture}
\caption{Visualizations of the MTRL policy on individual task performance.
\ral{
(a) The MTRL policy successfully stabilizes the quadrotor in a hover within seconds, even from high-speed tasks and challenging initial conditions.
(b) The MTRL policy successfully tracks the commanded velocity, even for challenging trajectories with speeds reaching up to 50 \si{\kilo\meter\per\hour}.
}
}
\vspace{-1em}
\label{fig:gradcam}
\end{figure*}

%% file: tables/realworld.tex
\begin{table}[t!]
    \centering
    \vspace*{2pt}
    \caption{Comparison of our MTRL policy's task performance between simulation and real world.}
    \label{tab:realworld}
    \setlength{\tabcolsep}{2pt}
    \begin{tabular}{cccc|cc|c}
    \toprule
     \multirow{2}*{\textbf{Methods}} 
     & \multicolumn{3}{c|}{\textbf{Racing}}
    &\multicolumn{2}{c|}{\textbf{Stabilization}}& \multicolumn{1}{c}{\textbf{Tracking}} \\
    &SR [\%] &Error [\SI{}{\meter}]  &  Time [\SI{}{\second}]
    &$t_{\text{fast}}$ [\SI{}{\second}]&$t_{\text{slow}}$ [\SI{}{\second}]&$\bm{e_v}$  [\SI{}{\meter\per\second}] \\
    
    \midrule
    \grayrow
    Simulation& 100\%& 0.152& 5.80 & 0.15 & 1.03 &0.99 \\
    Real-world& 100\%& 0.189& 5.94 & 0.17 & 1.19 &0.95 \\
    \bottomrule
\end{tabular}
\end{table}

%% file: sections/discusssion.tex
\subsection{Discussion}
Our experiments thoroughly evaluated the proposed MTRL framework, addressing its critical performance aspects.
First, we demonstrated that the MTRL approach significantly improves sample efficiency and task performance compared to single-task RL baselines. After training for half the amount of total samples (20M), our MTRL method reduced stabilization time by 18\% ($t_\text{half}$) and gate passing error (MGE) by 16\% compared even to individual RL methods.
Furthermore, even when trained to convergence, our MTRL approach consistently performs better than, or as well as other baseline methods.
Notably, it achieved a 6.7\% reduction in gate passing error in racing tasks compared to the next best approach.
In contrast, the MTRL-Separate baseline without the integration of shared and task-specific information resulted in a 0\% success rate in the racing task, as it could not even complete the task.
This failure underscores the importance of properly integrating shared and task-specific elements within our MTRL framework. 
Finally, the deployment of our MTRL policy in real-world scenarios confirmed its robustness, with performance closely matching that observed in simulations.

%% file: sections/conclusions.tex
\section{Conclusions}
In this work, we introduced the first Multi-Task Reinforcement Learning (MTRL) framework specifically designed for quadrotor control, addressing the challenges posed by diverse task requirements in real-world scenarios. 
By leveraging the shared physical dynamics of the quadrotor and employing a novel multi-critic setup with a shared task-agnostic observation encoder, our approach successfully integrates information across different tasks while maintaining high performance. 
The experimental results, both in simulation and real-world applications, demonstrated the effectiveness and efficiency of our method, particularly in enhancing sample efficiency when learning different tasks while maintaining strong task performance. 
This advancement paves the way for more versatile quadrotor control systems capable of performing a wide range of tasks within a single mission, thereby significantly contributing to the broader application of quadrotors in critical areas like search and rescue and infrastructure inspection.